\documentclass{ecai}
\usepackage{times}
\usepackage{graphicx}
\usepackage{latexsym}
\usepackage{url}
\usepackage{hyperref}
\usepackage{arydshln}
\usepackage{xcolor}
\usepackage{float}

\ecaisubmission   % inserts page numbers. Use only for submission of paper.
\begin{document}
% [X] Añadir ejemplo en OOV
% [X] Decir que va muy rápido y se pueden hacer muchos embeddings
% [X] Insistir un poco más en que es simple y raṕido y hemos clarificado todo el proceso de combinación
% [X] Separarse más de MEEMI 
% [X] MVM4 quitar de la tabla y poner en texto

\title{A Common Semantic Space for Monolingual \\ and Cross-Lingual Meta-Embeddings}

\author{Iker García-Ferrero \and Rodrigo Agerri \and German Rigau \institute{HiTZ Center - Ixa, University of the Basque Country UPV/EHU; \{iker.garciaf, rodrigo.agerri, german.rigau\}@ehu.eus }}

\maketitle
\bibliographystyle{ecai}

\begin{abstract}
This paper presents a new technique for creating monolingual and cross-lingual meta-embeddings. Our method integrates multiple word embeddings created from complementary techniques, textual sources, knowledge bases and languages. Existing word vectors are projected to a common semantic space using linear transformations and averaging. With our method the resulting meta-embeddings maintain the dimensionality of the original embeddings without losing information while dealing with the out-of-vocabulary problem. An extensive empirical evaluation demonstrates the effectiveness of our technique with respect to previous work on various intrinsic and extrinsic multilingual evaluations, obtaining competitive results for Semantic Textual Similarity and state-of-the-art performance for word similarity and POS tagging (English and Spanish). The resulting cross-lingual meta-embeddings also exhibit excellent cross-lingual transfer learning capabilities. In other words, we can leverage pre-trained source embeddings from a resource-rich language in order to improve the word representations for under-resourced languages.

\end{abstract}

\section{Introduction}

Word embeddings successfully capture lexical semantic information about words based on co-occurrence patterns extracted from large corpora \cite{mikolov2013efficient,pennington2014glove,mikolov2018advances} or knowledge bases \cite{bordes2011learning}, with excellent results on several tasks, including word similarity \cite{collobert2008unified,turian2010word,socher2011semi}, Semantic Textual Similarity \cite{shao2017hcti} or Sequence Labelling \cite{akbik2018contextual}.

Following the hypothesis that different knowledge sources may contain complementary semantic information \cite{goikoetxea2016single}, several authors have tried to enhance the quality of word representations by incorporating information from knowledge bases \cite{halawi2012large,faruqui2015retrofitting,bollegala2016joint,speer2017conceptnet}, other languages
\cite{vulic2016role,artetxe2018robust}, or both
%\cite{vulic2016role,artetxe2017learning,artetxe2018robust} or both 
\cite{speer2017semeval,Goikoetxea2018bilingual}.

Meta-embeddings  \cite{yin-schutze:2016:P16-1} aim to obtain an ensemble of distinct word embeddings each trained using different methods and resources, and possibly containing complementary information, to produce a word representation with an improved overall quality. Despite the potential benefits, learning a single meta-embedding from multiple source embeddings remains a challenging task mostly due to the fact that it is not obvious how to combine embeddings that are the result of using different algorithms, resources and languages. %In fact, previous approaches based on concatenation \cite{goikoetxea2016single}, averaging \cite{coates2018frustratingly} or dimensionality reduction \cite{raunak2017simple} produce meta-embeddings that suffer from a much larger dimensionality or from loss of information.

This paper explores the combination of multiple word embeddings (independently learned applying various techniques to different textual sources, knowledge bases and languages) by projecting them to a common semantic space using linear transformations, by generating appropriate representations for the out-of-vocabulary words, and by averaging.

%\color{red}
%NOTA: Creo que aquí deberíamos contar más sobre que ocurre con los meta embeddings anteriores y por que los nuestros son mejores. Es decir, toda la historia de que se pierde información, que hay que tratar el OOV...

The main challenge when generating meta embeddings is preserving the information encoded in the source embeddings. Concatenation \cite{goikoetxea2016single} has been used as a very strong baseline in all the previous works because the resulting meta-embedding preserves all the information from the source embeddings. However, concatenation generates meta-embeddings with a very high dimensionality. To avoid this, other methods have been proposed:  dimensionality reduction \cite{raunak2017simple}, averaging \cite{coates2018frustratingly} or complex linear methods and supervised neural models \cite{bollegala2018think,bollegala-bao-2018-learning,yin-schutze:2016:P16-1}. However, we show that previous methods cause some information loss that results in worse performance compared to concatenation. Interestingly, Coates and Bollegala \cite{coates2018frustratingly} mathematically proved that averaging is a good approximation to concatenation when the embeddings are placed in the same space. Following this, we use VecMap \cite{artetxe2016emnlp,artetxe2018aaai} to project the source word embeddings to the same semantic space. This allows us to minimize the lose of information when finally averaging the word embeddings. 

We also empirically demonstrate the benefits of treating the OOV words (when a word is missing in one of the source embeddings) problem. Simple strategies such as assigning a vector of zeros for OOV words have a large negative impact because these words are projected to very different positions than the ones expected. Thus, in this paper we propose a method to deal with the OOV problem which is suitable for mono- and cross-lingual meta-embedding generation.

Previous proposed methods to generate meta-embeddings also face the problem of hyper-parameter tuning, namely, to the fact that hyper-parameters need to be tuned depending on the source embeddings \cite{bollegala2018think,bollegala-bao-2018-learning,yin-schutze:2016:P16-1}. This means that the performance can vary significantly depending on the hyper-parameters used, which makes these methods difficult to use. As opposed to this, our approach does not rely on hyper-parameter tuning. In fact, we only use one hyperparameter which is kept constant for all our experiments. While making our method much robust, this also allows to successfully combine a large variety of word embeddings quickly and efficiently.

The main contributions of this paper are the following. First, we present a new method to generate meta-embeddings that exploits existing vector space mapping techniques. Second, we extensively demonstrate that treating correctly the OOV words problem (when a word is missing in one of the source embeddings) improves performance on both intrinsic and extrinsic evaluations. Third, our new approach to deal with the OOV problem works well for the generation of both mono- and cross-lingual meta-embeddings. Thus, while previous methods are only suitable for combining monolingual embeddings or require embeddings from the same source in order to achieve good results \cite{bollegala2018think,bollegala-bao-2018-learning,yin-schutze:2016:P16-1}, our approach allows to effectively and efficiently ensemble any number of existing mono- and cross-lingual embeddings. Fourth, we evaluate our embeddings in both intrinsic and extrinsic tasks, showing that our approach generates better static meta-embeddings than current state-of-the-art methods, including other linear methods and supervised neural models. In particular, we report new state-of-the-art results for Word Similarity and for POS tagging (UD 1.2) in English and Spanish. Moreover, the results show that the resulting cross-lingual meta-embeddings also exhibit excellent cross-lingual transfer learning capabilities. Finally, our code and meta-embeddings are publicly available\footnote{\url{https://github.com/ikergarcia1996/MVM-Embeddings}}.

% Our approach is able to leverage efficiently state-of-the-art corpus-based word representations \cite{mikolov2013efficient,pennington2014glove,mikolov2018advances}, knowledge-based word representations \cite{goikoetxea2016single,speer2017conceptnet} and cross-lingual representations \cite{speer2017conceptnet} without increasing the dimensionality of the resulting meta-embedding and without losing information in the process. We tested our monolingual and cross-lingual meta-embeddings in both intrinsic (word similarity) and extrinsic tasks (Semantic Textual Similarity, Part-of-Speech tagging and Named Entity Recognition). Our meta-embeddings improve other alternative state-of-the art meta-embedding approaches for all four tasks. Our code and meta-embeddings are publicly available\footnote{Anonymous github repository}.
%\color{}

The remainder of the paper is as follows. Section \ref{related-work} presents some alternative meta-embedding approaches. Section \ref{our-method} describes our meta-embedding creation approach and in Section \ref{Resources} we present the pre-trained word embeddings used in the experiments reported in Section \ref{experiments}. Finally, Section \ref{concusions} concludes and discusses future work.

\section{Related Work}
\label{related-work}

Previous research in word embedding learning methods have shown that word embeddings learnt using different methods and resources present significant variation in quality. For instance, Hill et al. \cite{hill2014not} show that word embeddings trained from monolingual or bilingual corpora capture different nearest neighbours. 

The term meta-embedding was coined by Yin and Sch\"utze \cite{yin-schutze:2016:P16-1}. They showed how to combine five different pre-trained word embeddings using a small neural network for improving the accuracy of cross-domain part-of-speech (POS) tagging. Following this research line, Bollegala et al. \cite{bollegala2018think} propose an unsupervised locally linear method for learning meta-embeddings from a given set of pre-trained source embeddings while Bollegala and Bao \cite{bollegala-bao-2018-learning} proposed three types of autoencoders for the purpose of learning meta-embeddings.

Although word embeddings are mainly constructed by exploiting information from text corpora only \cite{mikolov2013efficient,pennington2014glove,mikolov2018advances}, some research tries to combine it with the knowledge encoded in lexical resources such as WordNet \cite{halawi2012large,bollegala2016joint,goikoetxea2016single}, PPDB \cite{faruqui2015retrofitting} or ConceptNet \cite{speer2017conceptnet}.
Goikoetxea et al. \cite{goikoetxea2016single} show that simply concatenating word embeddings derived from text and WordNet outperform alternative methods such as retrofitting \cite{faruqui2015retrofitting} at the cost of increasing the dimensionality of the meta-embeddings. Coates and Bollegala \cite{coates2018frustratingly} prove that averaging is in some cases better than concatenation, with the additional benefit of a reduced dimensionality.
The most popular approach to address the dimensionality problem is to apply dimensionality reduction algorithms such as SVD \cite{yin-schutze:2016:P16-1}, PC \cite{ghannay2016word} or DRA \cite{raunak2017simple}. In this line of work, Numberbatch \cite{speer2017conceptnet} claims to be the best meta-embedding model so far. Numberbatch combines the knowledge encoded in ConcepNet, word2vec, GloVe and OpenSubtitles corpus using concatenation, dimensionality reduction and a variation on retrofitting.

Cross-lingual word embeddings also combine the knowledge from different languages and can be created using parallel corpora \cite{gouws2015bilbowa} or some sort of bilingual signal. An alternative approach is to train the embeddings for each language independently on monolingual corpora, and then map them to a shared common space based on a bilingual dictionary \cite{mikolov2013exploiting}. This requires minimal bilingual supervision compared to other approaches, while still leveraging large amounts of monolingual corpora with very competitive results \cite{artetxe2016emnlp,artetxe2018robust}.

The work from Doval et al. \cite{doval:meemiemnlp2018} is the most similar to ours.
However, instead of creating monolingual and cross-lingual meta-embeddings, their goal is to improve the multilingual mappings from VecMap \cite{artetxe2018robust} and MUSE \cite{conneau2017word} by averaging the two source embeddings once they are mapped into a common vector space. After the mapping, they use a linear mapping to predict the averages for OOV words. Their method is evaluated in word similarity, bilingual dictionary induction and cross-lingual hypernym discovery. Our approach differs in three significant ways: First, using our method we can combine any number of monolingual and cross-lingual source embeddings in any language. Second, we propose a new algorithm to generate representations for those words missing in some of the source embeddings. This new algorithm is compatible with the generation of both mono- and cross-lingual meta-embeddings thereby allowing us to improve the cross-lingual mapping and the quality of the final meta-embeddings. This means that we can apply transfer learning to exploit pre-trained source embeddings from a resource-rich language in order to improve the word representations for an less-resourced language. Finally, we also carried out extensive intrinsic (multilingual and cross-lingual word similarity) and extrinsic evaluations (multilingual Semantic Textual Similarity and multilingual POS tagging). 
%This evaluation also "allow us" to explain 

%explain why using our procedure (vector space alignment followed by averaging) produces better meta-embeddings that other approaches. 

%In this sense, the intrinsic and extrinsic evaluations performed (monolingual and cross-lingual) show that our method improves results for both types of tasks. SHOULD WE MENTION SOTA RESULTS FOR SOME OF THEM?? Finally, this means that we can use transfer learning to exploit pre-trained source embeddings from a resource-rich language in order to improve the word representations for an under-resourced language.
%In our research we also explain why using this procedure (vector space alignment followed by averaging) produces better meta-embeddings that other approaches. 

%The cross-lingual meta-embeddings with new, automatically generated, representations for missing words improve the quality of the meta-embeddings generated.

\section{A common cross-lingual semantic space}
\label{our-method}

Our approach to generate meta-embeddings consists of three steps: (i) aligning the vector spaces using VecMap, (ii) creating new representations for the missing words in the source embeddings and, (iii) averaging the resulting word embeddings. Our method can combine any number of word embeddings generated with any technique, source or language as long as there is some common vocabulary between them. Thus, the resulting meta-embedding vocabulary will be the union of the vocabularies of the word embeddings used. Henceforth, we will be referring to our approach as MVM (Meta-VecMap).

\subsection{VecMap}
\label{vecmap}

Artetxe et al. \cite{artetxe2016emnlp} presents a method called VecMap that learns bilingual mappings between word embeddings. VecMap first applies a normalization to the word embeddings and then an orthogonal transformation. Orthogonality allows monolingual invariance during the mapping, preserving vector dot products between words. Monolingual invariance \ ensures that no information loss will occur during the mapping step, which is desirable for our aim of generating meta-embeddings. As recommended by the authors of VecMap, we first apply length normalization and mean centering to the source word embeddings  (Step 0 as described in \cite{artetxe2018aaai}).  Then we apply the orthogonal mapping (Step 2). In this way, we are able to sequentially project to a common space any number of word embeddings. In fact, we can use the vector space of any of the source embeddings as a target common space to map the rest of the source embeddings. Finally, we generate the mapping dictionaries for VecMap as the intersection of the vocabularies of the source word embeddings to be ensembled.

Recent versions of VecMap introduce additional steps to improve bilingual mappings \cite{artetxe2018aaai}, but they are not relevant for our objective of generating meta-embeddings. Finally, it should be noted that VecMap cannot be used directly to generate meta-embeddings because it outputs two distinct word embeddings into the same space, not a meta-embedding.

\subsection{OOV treatment} \label{OVVgeneration}

\begin{figure}
\begin{center}
    \includegraphics[width=6cm]{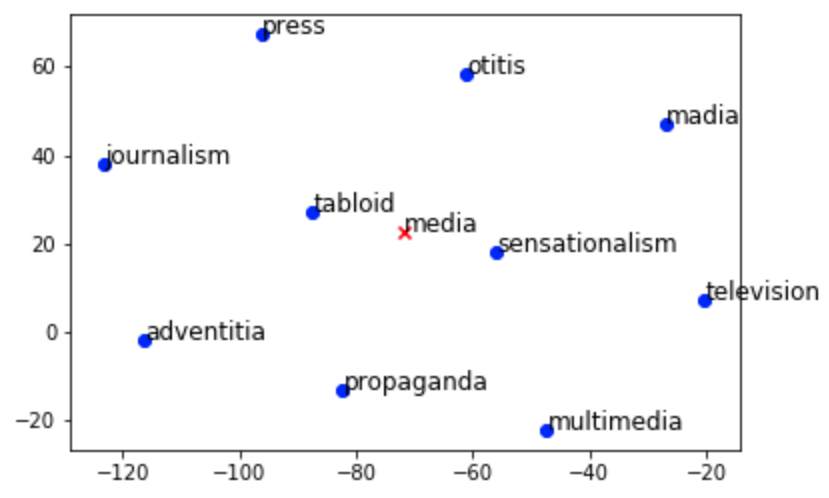}
\end{center}
\caption{Example of the NN approach for the word ``media''. Image generated using t-SNE}
\label{OOV_picture}

\end{figure}

When combining word embeddings we need to deal with the problem of having a missing word representations in one of the source embeddings. In this case, one source embedding E1 has a representation for the word W while another source embedding E2 does not have a representation for that word. Inspired by Speer and Lowry-Duda \cite{speer2017semeval} we also calculate a synthetic representation for the missing words. However, instead of using the neighbours in ConceptNet, once placed in a common space, we approximate a synthetic representation for a missing word (W) in one source embedding (E2) by using the ten most similar word representations of W in the other source embedding (E1). That is, first we calculate the set of ten nearest neighbours of W by cosine similarity in E1. Second, we obtain the corresponding embeddings in E2 of the set of nearest neighbour words. Finally, we assign the centroid in E2 of this set of nearest neighbours as the new word representation for W in E2. In this way we ensure that both source representations have the same vocabulary placed in a common space. In this paper, we refer to this approach as \textbf{NN} (Nearest Neighbour).
\autoref{OOV_picture} shows a real example of NN for two word embeddings (UKB and FT, both described in section \ref{Resources}). FT includes a presentation for the word ``media'' while UKB does not. Following our approach, first we find the ten nearest neighbours to ``media'' in the space of FT that are also present in UKB. These words are: ``multimedia'', ``press'', ``madia'', ``propaganda'', ``journalism'', ``otitis'', ``tabloid'', ``sensationalism'', ``television'' and ``adventitia''. Then, we create a new synthetic representation for ``media'' as the centroid of the nearest neighbour representations in the UKB space. 

\subsection{Averaging}
\label{average}

Finally, once projected two source word embeddings to a common space, we still need to combine them into a single word representation. The simplest way is by averaging the two projected word representations.

%Preliminary experiments on word similarity showed that alternative approaches such as not creating new representations for OOV or applying a different combination method instead of averaging, were detrimental for performance.

It should be noted that when generating cross-lingual meta-embeddings we also apply exactly the same process. That is, projection (\ref{vecmap}), OOV generation (\ref{OVVgeneration}) and averaging (\ref{average}).
%For instance, when creating a new meta-embedding for English and Spanish, at least one of the embeddings used must be cross-lingual in those two languages.
After projecting both embeddings to the same common space using VecMap, for every missing word in both source embeddings, we first apply the same OOV generation process described above (NN method) and, finally, we apply averaging. For example, when combining a monolingual Spanish embedding and a cross-lingual Spanish-English one, we will also create new synthetic embeddings for all missing words in the corresponding sets (possibly new English word embeddings in the Spanish set and possibly some new Spanish word embeddings in the cross-lingual set).
Thanks to this, our MVM (Meta-VecMap) meta-embedding approach can help to improve both languages. This is specially interesting when good representations are available only for one language.

\section{Word Embeddings}
\label{Resources}
In this section we describe the pre-trained word embeddings used to generate our monolingual and cross-lingual meta-embeddings. We also include a description of alternative state-of-the-art meta-embeddings.

 %implements a predictive model with two neural models: CBOW and Skip-gram. CBOW tries to predict a word given its context, and skip-gram the context of a given word.
We use the \textbf{Word2Vec} (W2V) \cite{mikolov2013efficient} embeddings from Google News (100 billion words). From \textbf{GloVe} (GV) \cite{pennington2014glove},
%is a count based model that uses the information of the concurrency between words and contexts.
the Common Crawl vectors (600 billion words). As recommended by the authors, we apply a ${l2}$ normalization to its variables. For \textbf{FastText} (FT) \cite{mikolov2018advances} %is an extension of word2vec that treats each word as the sum of its composing characters (ngrams)
in English we also use the Common Crawl (600 billion words) while for Spanish we use the vectors trained on the Spanish Billion Word Corpus (1.4 billion words)\footnote{Cristian Cardellino: https://crscardellino.github.io/SBWCE/}.

%\textcolor{red}{WE DO NOT COMPARE OR DESCRIBE MEETING IN THE MIDDLE, WHY?}

\textbf{RWSGwn} (UKB) \cite{goikoetxea2015random} combines random walks over WordNet with the skip-gram model.
We have used the vectors trained using WordNet3.0 plus gloss relations. Apparently using GloVe instead of skip-gram improves the performance of the final embeddings, so by default GloVe is used.

\textbf{Attract Repel} (AR) \cite{mrkvsic2017semantic} improves word embeddings by injecting synonymy and antonym constraints extracted from monolingual and cross-lingual lexical resources.
We used the English vocabulary from the four-lingual (English, German, Italian, Russian) vector space.

\textbf{Paragram} (P) \cite{wieting2015paraphrase} pre-trained word vectors learned using word paraphrase pairs from PPDB \cite{ganitkevitch2013ppdb} using a modification of the skip-gram objective function.
The hyper parameters were tuned using the wordsim-353 dataset. The word embeddings of the default model are initialized with glove word vectors.

\textbf{Numberbatch} (N) \cite{speer2017conceptnet} combines knowledge encoded in ConcepNet, word2vec, GloVe and OpenSubtitles 2016 using concatenation, dimensionality reduction and a variation on retrofitting. Numberbatch version 17.06 is used.

\textbf{JOINTChyb} (J) \cite{Goikoetxea2018bilingual} combines Random Walks over multilingual WordNets and bilingual corpora as input for a modified skip-gram model that forces equivalent terms in different languages to come closer during training.
We used the English-Spanish bilingual embeddings publicly available.

Other word embeddings such as LexVec \cite{salle2016matrix}, PDC/HDC \cite{Fei:Learning} or context2vec \cite{melamud2016context2vec} showed no significant improvements over the previous ones.

\section{Experiments}\label{experiments}

We tested our meta-embeddings in both intrinsic (word similarity) and extrinsic tasks, namely, Semantic Textual Similarity, POS tagging and Named Entity Recognition.

\subsection{Word Similarity}

First we compare our meta-embeddings against the most important baseline methods (concatenation, dimensionality reduction and average) and alternative meta-embeddings for English. Furthermore, we also evaluate our MVM meta-embeddings in a multilingual and cross lingual setting.
For testing we have used the fourteen datasets described in the section 4.1.1 of Bakarov \cite{bakarov2018survey}.
The similarity between two words has been calculated as the cosine between their word representations. The correlation between the similarities calculated using the word embeddings and the gold scores provided by the dataset has been calculated using Spearman correlation.% When evaluating word embeddings in a word similarity dataset we usually face the Out-of-Vocabulary problem (OOV), since the embedding may not have a representation for every word in the dataset.

\begin{table}
\centering
\small
\begin{tabular}{llll}
Embedding &  Cov. All  & Cov. Rel & Cov. Sim \\
\hline
  W2V & 97.4 &	98.3 &	97.3 \\
  GV & 99.8	& 100 & 	99.7 \\
  FT & 99.8	& 100	& 99.7 \\
  UKB & 89.0 &	94.4 &	87.3 \\
  AR & 94.2	& 99.7	& 92.1 \\
  P & 91.8	& 97.1 &	90.2 \\
%  LV & 99.6 &	100.0 &	99.5 \\
\end{tabular}
\caption{Coverage in the word similarity datasets}
\label{Coverage}
\end{table}

Table \ref{Coverage} shows the coverage of the word embeddings in the word similarity datasets. UKB presents the lower coverage while GloVe and FastText exhibit almost full coverage. Multiple approaches have been applied to deal with the Out-of-Vocabulary problem (OOV). For instance, in order to assign an arbitrary vector for a missing word, previous works assign a predefined similarity (i.e. 0.5) or, alternatively, do not take into account the pairs of words when there is no representation for at least one of the embeddings. For a fairer comparison, we use the latter option since assigning arbitrary values cause embeddings with larger vocabularies (i.e. our own meta-embeddings) to obtain better results.

%Assigning arbitrary values cause embeddings with larger vocabularies to obtain better results. Since the vocabulary of our meta-embeddings is the union of the vocabularies of the source embeddings, they would score higher results using this approach even if the quality of the representation does not increase. Therefore, we adopt the last option to handle OOV words.

 %We adopt this last option. As we mentioned earlier, the vocabulary of one of our meta-embeddings is the union of the vocabularies of the individual embeddings that form th meta-embedding.

\subsubsection{English Word Similarity}

In this section we first compare our meta-embeddings against some baseline ensemble methods, and then compare our method against other methods and meta-embeddings. For this comparison we used all the datasets described by  Bakarov \cite{bakarov2018survey}. For brevity, we just report the average of the results of all the datasets (Av), the average of those datasets that measure word similarity (Sim), and the average of the ones that measure word relatedness (Rel). The WS353 dataset is divided in two subsets \cite{agirre2009study}. In this section all the meta-embeddings have been mapped to the vector space of the English FastText (Common Crawl, 600B tokens).

The main three baseline methods to compare our approach with are the following. \textbf{Concatenation}: For each word we concatenate the representation for that word in each source embedding. To ensure that every source embedding contributes equally in the meta-embedding we normalize them using the ${l_2}$ norm. This method produces meta-embeddings of high quality but with a very high dimensionality. \textbf{Dimensionality reduction}: The most popular approach to solve the dimensionality problem is to apply dimensionality reduction algorithms. We report the results obtained using DRA \cite{raunak2017simple} when reducing the dimensionality to 300 dimensions of the meta-embeddings generated by concatenation. We also tested the implementations of PCA and truncated SVD using sklearn \cite{scikit-learn} but DRA obtains the best results. 
\textbf{Averaging}: For each word, we obtain the average of the representation for that word in each source embedding. To ensure that every source embedding contributes equally, we normalize using the ${l_2}$ norm. %This is another way to approximate the results of the concatenation keeping the number of dimensions low
Coates and Bollegala \cite{coates2018frustratingly} probe that averaging is a way to approximate the results of the concatenation keeping the number of dimensions low. 
%We also apply the OOV approach described in Section \ref{OVVgeneration} when concatenating or averaging word embeddings.

Before proceeding with the Word Similarity experiments, we first we tested our OOV generation algorithm. Table \ref{tab:OOV} reports the results of our meta-embeddings combining the embedding with larger (FastText) and smaller  (UKB) coverage. We present the results in the two word similarity datasets where UKB has smaller coverage (Mturk287 and Rare Words). In the case of concatenation, we compare our OOV generation algorithm (NN) with the most common approach appearing in the literature: using a vector of zeros. Regarding average, we compare our approach against averaging using the representations available. As we can observe, our OOV generation algorithm significantly improves the results obtained by averaging and are close or surpass the ones obtained by concatenation. Interestingly, the mapping using VecMap by itself is a good method for OOV handling, increasing by a large margin the results of the concatenation using zeros and the average without mapping. We obtain even a better approximation to the results of concatenation by combining the mapping method (MVM) and our word generation algorithm (NN).

\begin{table}
\centering
\small
\begin{tabular}{lll}
Embedding &  Mturk-287 & RW \\    
\hline
\multicolumn{3}{l}{\textbf{Source embeddings}} \\

FastText & 72.6 (100) & \textbf{59.5} (98.1)\\
UKB & 64.6 (80.1) & 45.0 (68.9)\\
\hline
\multicolumn{3}{l}{\textbf{Baseline OOV treatment}} \\

Concatenation & 71.5 (100) & 51.4 (99.9) \\
Average & 66.2 (100) & 50.2  (99.9)\\
MVM & 71.6 (100) & 54.5 (99.9)\\

\hline
\multicolumn{3}{l}{\textbf{Using NN}} \\
Concatenation & \textbf{72.7} (100) & 55.9 (99.9) \\
Average  & 68.1 (100) & 54.8 (99.9)\\
MVM  & 71.8 (100) & 58.7 (99.9)\\

\end{tabular}

\caption{Results of the different OOV methods on the meta-embedding generation in the Mturk-287 and RareWords (RW) word similarity datasets. The coverage of the embeddings for each dataset is indicated between brackets after the results. In bold the best results overall.}
\label{tab:OOV}
\end{table}

\begin{table}
\centering
\small
\begin{tabular}{llll}
    & Av & Rel & Sim \\
  \hline
  \textbf{Source Embeddings} \\
  W2V & 60.7 & 68.7 & 57.9 \\
  GV & 62.4 & 75.3 & 57.8 \\
%  LV & 62.6 & 73.4 & 58.7 \\
  UKB & 62.5 & 70.2 & 60.3 \\
  AR & 65.0 & 61.4 & \bf{67.0} \\
  FT & 66.6 & \bf{76.5} & 63.2 \\
  P & \bf{68.2} & 74.9 & 65.9 \\
  \hline
  \textbf{Concatenation} \\

  W2V+UKB & 68.6 & 75.8 & 66.4 \\
  W2V+AR & 69.1 & 71.4 & 68.6 \\
  GV+UKB & 69.4 & 79.0 & 66.3 \\
  GV+AR & 70.1 & 75.4 & 68.5 \\
  FT+W2V & 65.1 & 74.5 & 61.8 \\
  FT+GV & 65.7 & 77.6  & 61.5 \\
  FT+UKB & 70.9 & 78.8 & 68.3 \\
  FT+AR & 71.5 & 74.6 & 70.8 \\
  FT+UKB+P & 72.8 & \bf{81.0} & 70.1 \\
  FT+UKB+AR & 72.7 & 78.2 & 71.1 \\
%  FT+UKB+AR+LV & 72.4 & 79.4 & 69.9 \\
  FT+UKB+AR+P & \bf{74.1} & 79.9 & \bf{72.5} \\
  \hline
 \textbf{Average} \\
 
  W2V+UKB & 66.8 & 72.4 & 64.9 \\
  W2V+AR & 67.1 & 68.6 & 67.1 \\
  GV+UKB & 67.8 & 76.6 & 65.0 \\
  GV+AR & 70.5 & 76.2 & 68.8 \\
  FT+W2V & 63.3 & 72.3 & 60.1 \\
  FT+GV & 63.6 & 75.0 & 59.6 \\
  FT+UKB & 67.1 & 74.8 & 64.9 \\
  FT+AR & 69.0 & 70.9 & 68.6 \\
  FT+UKB+P & 68.9 & \bf{77.0} & 66.5 \\
  FT+UKB+AR & 69.6 & 72.6 & 69.2 \\
  %FT+UKB+AR+LV & 69.2 & 74.1 & 67.7 \\
  FT+UKB+AR+P & \bf{71.6} & 75.9 & \bf{70.5} \\
  \hline
  \textbf{Dim. reduction (DRA)}  \\
  
  W2V+UKB  & 64.5  & 69.9 & 62.7 \\
  W2V+AR & 67.2 & 67.6 & 67.0 \\
  GV+UKB  & 68.0 & 77.8 & 64.6 \\
  GV+AR  & 70.9 & 77.0 & 68.7 \\
  FT+W2V  & 65.1 & 73.9 & 61.8 \\
  FT+GV  & 66.0 & 77.5 & 61.8 \\
  FT+UKB & 68.9 & 76.9 & 66.3 \\
  FT+AR  & 72.1 & 75.1 & 70.9 \\
  FT+UKB+P & 71.3 & \bf{78.5} & 69.0 \\
  FT+UKB+AR  & 71.9 & 76.1 & 70.6 \\
%  FT+UKB+AR+LV  & 72.5 & 78.4 & 70.6 \\
  FT+UKB+AR+P  & \bf{72.8} & 77.3 & \bf{71.4} \\
  \hline
  \textbf{MVM} \\
  
  W2V+UKB & 68.7 & 77.0 & 66.2 \\
  W2V+AR & 69.4 & 72.8 & 68.4 \\
  GV+UKB & 69.4 & 78.6 & 66.4 \\
  GV+AR & 70.6 & 75.5 & 69.1 \\
  FT+W2V & 65.2 & 74.9 & 61.8 \\
  FT+GV & 66.4 & 77.7 & 62.3 \\
  FT+UKB & 71.8 & 79.4 & 69.6 \\
  FT+AR & 72.1 & 75.4 & 71.2 \\
  FT+UKB+P & 73.2 & \bf{81.0} & 70.7 \\
  FT+UKB+AR & 74.0 & 78.6 & \textbf{72.9} \\
%  FT+UKB+AR+LV & 73.5 & 79.8 & 71.6 \\
  FT+UKB+AR+P & \bf{74.5} & 80.0 & \bf{72.9} \\

\end{tabular}

\caption{English Word Similarity results}
\label{Resultadosgeneral}
\end{table}

Table \ref{Resultadosgeneral} report the results of the pre-trained embeddings, the baseline meta-embeddings and our meta-embeddings. All these meta-embeddings also use our OOV approach (NN). In general, the concatenation scores are significantly better than the ones of the best pre-trained word embeddings used in the concatenation. Concatenating pre-trained embeddings outperform meta-embeddings created by averaging or by dimensionality reduction. Dimensionality reduction obtains slightly better results than averaging, but at the cost of much more computational effort.

The results in Table \ref{Resultadosgeneral} show that our approach (MVM) produces meta-embeddings of similar or higher quality as the ones produced by concatenation, with the advantage of not increasing the dimensionality of the word vectors.

In general, meta-embeddings generate better word representations when ensembling pre-trained word embeddings that encode complementary and different knowledge. For instance, the combination of word representations learned from WordNet and text results in higher performance \cite{goikoetxea2016single}.

We also test our meta-embeddings with respect to alternative approaches. \textbf{Autoencoding Meta-Embeddings (AAEME)} \cite{bollegala-bao-2018-learning} applies autoencoders to generate meta-embeddings. Autoencoders are an unsupervised learning method that first compress the input in a space of latent variables and then reconstructs the input based on the information encoded in these latent variables. This method aims to learn meta-embeddings by reconstructing multiple source embeddings. This method comes in three flavours, DAEME, CAEME and AAEME. We used the last one because it obtains the best results. We applied default parameters and enabled the option to generate OOV word representations. We also compare with the \textbf{Locally Linear Meta-Embedding Learning} (LLE) \cite{bollegala2018think} approach which consists of two steps. In the reconstruction step the embeddings of each word are represented by the linear weighted combination of the embeddings of its nearest neighbours. In the projection step the meta-embedding of each word is computed such that the nearest neighbours in the source embedding spaces are embedded closely to each other in the meta-embedding space. We tested this method with the same parameters used in the original paper.

We also tested several configurations of contextual embeddings such as those from ELMo \cite{Peters:2018} or BERT \cite{devlin2019bert} but as this task does not include context, they obtain much lower results than the static ones.

\begin{table}
\centering
\small
\begin{tabular}{llll}
Embeddings & Av & Rel & Sim \\
\hline
\textbf{Numberbatch} \\

 N & 73.7 & 78.9 & 71.6 \\
\hline
\textbf{AAEME} \\

FT+UKB+P & 65.8 & 74.3 & 62.7 \\
FT+UKB+AR+P & 65.0 & 73.6 & 61.9 \\
\hline
\textbf{LLE }\\

FT+UKB+P  & 63.2 & 71.5 & 60.0 \\
FT+UKB+AR+P  & 64.9 & 71.4 & 62.3 \\
\hline
\textbf{MVM }\\

FT+UKB+P & 73.2 & 81.0 & 70.7 \\
FT+UKB+AR+P & 74.5 & 80.0 & 72.9 \\
\hdashline
FT+UKB+N & 74.1 & \textbf{81.3} & 71.8 \\
FT+UKB+AR+N & \textbf{75.4} & 80.4 & \textbf{73.9} \\

\end{tabular}

\caption{Comparison of different meta-embeddings in the English Word Similarity task}
\label{ResultadosvsOtrosMetodos}
\end{table}

Table \ref{ResultadosvsOtrosMetodos} presents the results comparing our meta-embeddings (MVM) with respect to different meta-embedding approaches (AAEME and LLE) and Numberbatch. First, our meta-embeddings obtain similar or better results than Numberbatch and outperforms by far LLE and AAEME  when using the same set of pre-trained embeddings.

The results show that Numberbatch encodes additional knowledge not included in the rest of pre-trained embeddings. Adding Numberbatch to our meta-embeddings results in even better meta-embeddings. To the best of our knowledge these are the best published results so far on the English word similarity task.

\subsubsection{Multilingual and Cross-lingual Word Similarity }

We also tested our meta-embedding generation method in multilingual and crosslingual word similarity tasks. We used English and Spanish since for these languages there are many pre-trained word embeddings and evaluation datasets. First, we evaluate our cross-lingual English-Spanish meta-embeddings in the English version of the SimLex999, WS353, RG65 and SemEval datasets, and then we evaluate the same meta-embeddings in the Spanish version of the datasets.
In the cross-lingual English-Spanish task we used the same cross-lingual meta-embeddings but this time for each pair in the datasets, the first word in English and the second in Spanish.

We found out that the best approach to generate cross-lingual meta-embeddings using our method is to independently generate the best possible meta-embeddings in English and the best possible meta-embeddings in Spanish, and project them to the same space.
Thus, we decided to map our meta-embeddings to the English-Spanish cross-lingual space offered by Numberbatch (N) or JOINTChyb (J). This way it is not necessary to use or induce dictionaries and the generation of the cross-lingual meta-embeddings is much faster. In particular, we decided to project our meta-embeddings to the space of JOINTChyb (J).

Our approach for creating meta-embeddings can ensemble a very large number of word embeddings. Table \ref{cross-lingual-embeddings} summarises the information of our best performing meta-embeddings. Subscripts (en) for English or (es) for Spanish indicate the language of the monolingual embeddings selected for creating the meta-embeddings. For instance, FT$_{en}$ corresponds to the English FastText. The subscript can also indicate the language of the words selected for creating the meta-embeddings. For instance, J$_{en}$ corresponds to the English vocabulary JOINTChyb. Thus, our meta-embedding MVM1$_{enes}$ includes 6 word embeddings, four English and two Spanish, all projected to the space of JOINTChyb which is the same space for both English and Spanish.

 \begin{table}
    \small
 \centering
     \begin{tabular}{lll}
          Name & EN & ES \\
          \hline
            MVM1$_{enes}$ & J$_{en}$+FT$_{en}$+AR$_{en}$+P$_{en}$ & J$_{es}$+FT$_{es}$ \\
            MVM2$_{enes}$ & J$_{en}$+N$_{en}$ & J$_{es}$+N$_{es}$ \\
            MVM3$_{enes}$ & J$_{en}$+N$_{en}$+FT$_{en}$+AR$_{en}$ & J$_{es}$+N$_{es}$ \\
            %MVM4$_{enes}$ & J$_{en}$+N$_{en}$+FT$_{en}$+AT$_{en}$+P$_{en}$ & J$_{es}$+FT$_{es}$ \\
     \end{tabular}
     \caption{Cross-lingual meta-embeddings}
     \label{cross-lingual-embeddings}
 \end{table}

Table \ref{cross-lingual-word-similarity} shows the multilingual results for English and Spanish, and the cross-lingual English-Spanish. We include the results of some state-of-the-art monolingual embeddings, and the cross-lingual embeddings such as JOINTChyb (J), Numberbatch (N) and the ones from  Doval et al. \cite{doval:meemiemnlp2018} (MEEMI$_{enes}$) in MUSE and VecMap flavours\footnote{\url{https://github.com/yeraidm/meemi}}. Compared to all these embeddings, our cross-lingual meta-embeddings obtain the best results for English, slightly better for Spanish and slightly lower than Numberbatch on the cross-lingual datasets. Surprisingly, our cross-lingual results are slightly lower than Numberbatch due to an increase in coverage (less OOV). Providing answers for the same word-pairs as Numberbatch, our meta-embedding MVM3$_{enes}$ obtains in the cross-lingual English-Spanish a result of 80.4 which indicates a small improvement of the cross-lingual meta-embedding.

 \begin{table}
    \small
     \centering
     \begin{tabular}{llll}
          Embedding & EN-EN & ES-ES & EN-ES\\
          \hline
          \multicolumn{4}{l}{\textbf{Mono-Lingual baselines}} \\
          
          UKB$_{en}$ & 71.9 & -  & -  \\
          FT$_{en}$ & 73.5 & - & - \\
          FT$_{es}$ & - &  60.2 & - \\
          P$_{en}$ & 73.6 & - & - \\
          AR$_{en}$ & 75.7 & - & - \\
          \hline
          \multicolumn{4}{l}{\textbf{Cross-Lingual baselines}} \\
          J$_{enes}$ & 73.5 & 68.7 & 71.7 \\
          N$_{enes}$ & 80.7 & 75.6 & \bf{80.0} \\
          MEEMI$_{enes}$ (MUSE) & 66.6 & 62.6 & 62.4 \\
          MEEMI$_{enes}$ (VecMap) & 66.3 & 62.6 & 63.9 \\
          \hline
          \multicolumn{4}{l}{\textbf{Cross-lingual Meta-Embeddings}} \\
          
          MVM1$_{enes}$ & 81.6 & 69.1 & 74.6 \\
%          ME2 & 82.4 & \bf{75.7} & 79.6 \\
          MVM2$_{enes}$ & 81.1 & \bf{75.7} & 79.5 \\
          MVM3$_{enes}$ & \bf{82.6} & \bf{75.7} & 79.6 \\
%          M5 & 82.4 & 84.1 & 78.6 \\
          
     \end{tabular}
     \caption{Multilingual and Cross-lingual word similarity results using cross-lingual meta-embeddings. }
     \label{cross-lingual-word-similarity}
 \end{table}

\subsection{Semantic Textual Similarity}

Semantic Textual similarity aims to calculate the degree of semantic similarity between two sentences. This task is a key component of many NLP systems and has received a lot of attention in recent years.
The STS benchmark \cite{cer2017semeval} is a set of multilingual and cross-lingual datasets used from 2012 to 2017 in SemEval. In order to evaluate our embeddings in these datasets we have used a well-known convolutional neural model \cite{shao2017hcti}. This system can be trained very fast without the need for large computing power while obtaining very good results in the task. The system achieved the 3rd place on SemEval 2017. We tested all embeddings and meta-embeddings using the same default parameters.

For English we used the training, development and test set provided by STSbenchmark. For Spanish we collected all the Spanish data provided in previous SemEval editions to create a new training and development dataset. 85\% of sentences were used for training and 15\% for development. We used as test the dataset provided in the SemEval 2017 STS task 1. In total the English dataset has 5749 pairs for training, 1500 as development set and 1379 as training set. The English dataset is almost five times larger than the Spanish one, which consists of 1295 pairs for training, 324 as development test and 250 as test set. We trained the model 8 times using the development data for fine tuning. The model and epoch that achieves the higher result in the development data is evaluated in the test set.

%\begin{table}[tbh]
%\centering
%\small
%\begin{tabular}{llll}
%& Train & Dev. & Test \\
%\hline
%English & 5749 & 1500 & 1379 \\
%Spanish & 1296 & 324 & 250 \\
%\end{tabular}
%\caption{Number of words for each dataset}
%\label{num_words_STS}
%\end{table}

\begin{table}
\centering
\small
\begin{tabular}{lll}
Embedding & Dev & Test \\
\hline
%\textbf{Baselines}\\

UKB$_{en}$ & 76.3 & 70.7 \\
FT$_{en}$ & 81.7 & 76.1 \\
GV$_{en}$ & 81.8 & 78.1 \\
AR$_{en}$ & 81.3 & 75.3 \\
P$_{en}$ & 82.4 & 78.8 \\
N$_{en}$ & 83.5 & 79.6 \\
\hline
\textbf{MVM}\\

%FT + UKB + N & \textbf{84.7} & 79.6 \\ No contribution
FT$_{en}$+UKB$_{en}$+AR$_{en}$+P$_{en}$ & 83.5 & 79.7 \\
FT$_{en}$+UKB$_{en}$+AR$_{en}$+N$_{en}$ & \textbf{84.0} & 80.1 \\
FT$_{en}$+N$_{en}$+AR$_{en}$+P$_{en}$ & 83.9 & \textbf{80.5} \\
\end{tabular}
\caption{STS results training and testing on English}
\label{STSEN}
\end{table}

\begin{table}
\centering
\small

\begin{tabular}{lll}

Embedding & Dev & Test \\
\hline
%\textbf{Baselines} \\

GV$_{es}$ & 76.6 & 73.7 \\
FT$_{es}$ & 80.4 & 75.1 \\
J$_{es}$ & 76.0 & 73.1 \\
N$_{es}$ & 76.3 & 73.8 \\
\hline
\textbf{MVM} \\

N$_{es}$+FT$_{es}$ & \textbf{81.8} & 75.3 \\
N$_{enes}$+FT$_{en}$+FT$_{es}$+P$_{en}$ & 81.0 & \textbf{78.4} \\
\end{tabular}
\caption{STS results training and testing on Spanish}
\label{STSES}
\end{table}

Tables \ref{STSEN} and \ref{STSES} show the results on English and  Spanish, respectively. Some of our meta-embeddings configurations achieve the best results in both languages. The original system presented at SemEval used the GloVe pretrained embeddings. Now, using our meta-embeddings the same system outperforms the original one by a large margin.
Compared to English, for Spanish the results are much lower in general (maybe partially due to the smaller size of the training set). The best results on the test data for Spanish are obtained when including into the meta-embeddings not just Spanish embeddings, but also the English ones. This suggests that using our meta-embeddings the knowledge encoded in one language can improve the quality of the knowledge encoded in another language.

\begin{table}
\centering
\small
\begin{tabular}{p{4cm}ll}
Embedding & Test EN & Test ES \\
\hline
%\multicolumn{2}{l}{\textbf{Baseline Cross-Lingual Embeddings}}\\

J$_{enes}$ & 73.8 & 69.3 \\
N$_{enes}$ & \textbf{79.6} & 81.8 \\
\hline
%\multicolumn{2}{l}{\textbf{Cross-Lingual Embeddings}} \\

MVM4$_{enes}$ & 78.8 & \textbf{82.8} \\

\end{tabular}
\caption{STS training on English and testing on Spanish.}
\label{STSENES}
\end{table}

In the last evaluation for STS, Table \ref{STSENES} shows the results when training the model in English and then, without seeing any Spanish example, testing the model in Spanish. That is, using the training parameters learned from English and applying them to Spanish and English. The meta-embedding MVM4$_{enes}$ has been obtained by ensembling J$_{enes}$, N$_{enes}$, FT$_{en}$, AT$_{en}$, P$_{en}$ and FT$_{es}$. It should be noted that, contrary to previous MVM1, MVM2 and MVM3 meta-embeddings used for word similarity (where we independently generate the best possible meta-embeddings in English and the best possible meta-embeddings in Spanish, and project them to the same space) MVM4 has been generated as a single meta-embedding using English, Spanish and English-Spanish source embeddings.
The results show that our meta-embeddings not only improve the STS results in both, English and Spanish. They also exhibit very robust cross-lingual transfer learning capabilities, allowing us to train a Spanish STS model with English data only. Apparently, these results suggest a very promising cross-lingual transfer learning research.
%With these experiment we do not seek to obtain the best overall results for STS task (in fact, contextualized word embeddings are at the top of the leadearboard), but validate our meta-embedding approach by comparing it against other static word embeddings and meta-embeddings.
Although current STS state-of-the art results are currently being obtained by contextual embeddings, these experiments confirm that our meta-embeddings outperform other existing static embeddings and meta-embeddings. Furthermore, since contextual embeddings require huge resources, our cross-lingual approach could be useful for under-resourced languages.

%us, our cross-lingual meta-embeddings exhibit a very promising a
%Thanks to this property, our meta-embedding can significantly improve the performance of the STS task for languages for which we have very few examples for training or even none.

\subsection{POS Tagging}

For our experiments on a extrinsic Sequence Labeling task such as Part-of-Speech (POS) tagging we chose to use the Flair system \cite{akbik2018contextual}.  Flair implements a recurrent neural network (RNN) architecture to represent documents, modelling text as a sequence of characters passed to the RNN which at each point in the sequence is trained to predict the next character \cite{akbik2018contextual}. Most importantly, Flair provides their own type of character-based contextual embedding models which represent words as sequences of characters in context. In order to obtain such representations, they use backward and forward Long Short Term Memory (LSTM) models to predict the next character for each character in the sequence. Flair contextual character-based word representations have been successfully applied to sequence labelling tasks obtaining best results for a number of public NER and POS benchmarks \cite{akbik2018contextual}, outperforming current successful approaches such as BERT and ELMO \cite{devlin2019bert,Peters:2018}.

For best results, the Flair authors recommend to combine their own Flair embeddings with additional static embeddings such as Glove or FastText. Thus, we follow their advice and use their system leveraging the pre-trained word embedding and meta-embedding models combined with the Flair character-based contextual embeddings for English and Spanish.

\begin{table}
\centering
\small
\begin{tabular}{lll}
Embedding & en & es \\
\hline
Flair & 96.54 & 97.15 \\
Flair + GV (news) & 96.86 &  97.23 \\
Flair + FT (news) & 96.76 & 97.24 \\
Flair + Numberbatch & 96.69 & 97.26\\
Flair + JOINTCHyb & 96.90 & 97.27 \\ \hline
\textbf{Baseline}\\
Heinzerling and Strube \cite{heinzerling-strube-2019-sequence} & 96.10  & 96.80 \\
\hline
\textbf{MVM} \\
Flair + J$_{enes}$+N$_{enes}$+FT$_{es}$+FT$_{en}$+AT$_{en}$+P$_{en}$ & \textbf{96.96}  & \textbf{97.35} \\
\end{tabular}
\caption{Word accuracy for Spanish and English on UD 1.2.}
\label{tab:pos}
\end{table}

In order to facilitate comparison with previous state-of-the-art methods, we experiment with the Universal Dependencies 1.2 data, which provides train, development and test partitions. We train the neural network off-the-shelf using the parameters used in Akbik et al. \cite{akbik-etal-2019-pooled}, tuning the system on the development data and using the test only for the final evaluation. Table \ref{tab:pos} shows first the results obtained by training the system with the Flair character-based contextual embeddings only. After that we combine the Flair embeddings with other static embeddings and meta-embeddings. For this section we chose to compare our work with those embeddings that have previously been most effective in combination with Flair, namely, Glove and FastText. Furthermore, we also compare with the best previous meta-embeddings: Numberbatch and JOINTCHyb. The Baseline result reports the best published results so far for each task and dataset. Finally, we show our meta-embedding (MVM) result.

In Table \ref{tab:pos} we can see that for POS tagging the best performing meta-embedding combines the cross-lingual embeddings JOINTChyb, Numberbatch (full vocabulary), both Spanish and English FastText, Attract Repel and Paragram. Our meta-embeddings help to improve results over previous meta-embeddings. Furthermore, our best results establish new state-of-the-art performance for POS tagging using UD 1.2 for Spanish and English.

Finally, as it was the case for STS, we would like to highlight that the cross-lingual character of our meta-embeddings helps also for monolingual tasks. Looking at the obtained results, we hypothesize that our NN algorithm for the treatment of OOV words has a beneficial effect specially for the creation of cross-lingual meta-embeddings.

\section{Conclusions and future work}
\label{concusions}

We have presented a new meta-embedding generation approach that can integrate efficiently and effectively multiple word embeddings derived from complementary techniques, textual sources, knowledge bases and languages. Using our method the resulting meta-embeddings maintain the dimensionality of the original embeddings without losing information while dealing with the out-of-vocabulary (OOV) problem. Our meta-embeddings help to improve performance over previous mono- and cross-lingual word meta-embeddings, obtaining excellent results on several tasks, including new state-of-the-art results for Word Similarity and POS tagging. Furthermore, the resulting cross-lingual meta-embeddings also exhibit excellent cross-lingual transfer learning capabilities. That is, we can exploit pre-trained source embeddings from a resource-rich language in order to improve the word representations for less-resourced languages. Moreover, our method does not rely on hyper-parameter tuning. Finally, our software can generate high quality meta-embeddings from a large amount of source embeddings quickly and efficiently.

Most previous work on meta-embeddings has focused on combining different word embeddings from similar sources, such as word those generated from similar textual corpora \cite{bollegala-bao-2018-learning,bollegala2018think,yin-schutze:2016:P16-1}. Our results show the importance of combining source embeddings that encode complementary knowledge.
We plan to investigate further this technique to improve performance for under-resourced languages by applying cross-lingual meta-embeddings. We leave also for future work the creation of cross-lingual embeddings just from monolingual ones.

%\ack We would like to thank the referees for their comments, which helped improve this paper considerably

\section*{Acknowledgments}

This work has been funded by the~Spanish Ministry of Science, Innovation and Universities under the project DeepReading (RTI2018-096846-B-C21) (MCIU/AEI/FEDER, UE) and by the BBVA Big Data 2018 ``BigKnowledge for Text Mining (BigKnowledge)'' project. The second author is funded by the Ramon y Cajal Fellowship RYC-2017-23647. We also acknowledge the~support of the NVIDIA Corporation with the~donation of a Titan V GPU used for this research.

\bibliography{ecai}

\begin{thebibliography}{10}

\bibitem{agirre2009study}
Eneko Agirre, Enrique Alfonseca, Keith Hall, Jana Kravalova, Marius
  Pa{\c{s}}ca, and Aitor Soroa, `A study on similarity and relatedness using
  distributional and wordnet-based approaches', in {\em NAACL 2009}, (2009).

\bibitem{akbik-etal-2019-pooled}
Alan Akbik, Tanja Bergmann, and Roland Vollgraf, `Pooled contextualized
  embeddings for named entity recognition', in {\em NAACL}, pp. 724--728,
  (2019).

\bibitem{akbik2018contextual}
Alan Akbik, Duncan Blythe, and Roland Vollgraf, `Contextual string embeddings
  for sequence labeling', in {\em COLING 2018}, (2018).

\bibitem{artetxe2016emnlp}
Mikel Artetxe, Gorka Labaka, and Eneko Agirre, `Learning principled bilingual
  mappings of word embeddings while preserving monolingual invariance', in {\em
  EMNLP}, (2016).

\bibitem{artetxe2018aaai}
Mikel Artetxe, Gorka Labaka, and Eneko Agirre, `Generalizing and improving
  bilingual word embedding mappings with a multi-step framework of linear
  transformations', in {\em AAAI 2018}, (2018).

\bibitem{artetxe2018robust}
Mikel Artetxe, Gorka Labaka, and Eneko Agirre, `A robust self-learning method
  for fully unsupervised cross-lingual mappings of word embeddings', in {\em
  ACL 2018}, (2018).

\bibitem{bakarov2018survey}
Amir Bakarov, `A survey of word embeddings evaluation methods', {\em CoRR
  2018}, (2018).

\bibitem{bollegala-bao-2018-learning}
Danushka Bollegala and Cong Bao, `Learning word meta-embeddings by
  autoencoding', in {\em COLING 2018}, (2018).

\bibitem{bollegala2018think}
Danushka Bollegala, Kohei Hayashi, and Ken-Ichi Kawarabayashi, `Think globally,
  embed locally: locally linear meta-embedding of words', in {\em IJCAI 2018},
  (2018).

\bibitem{bollegala2016joint}
Danushka Bollegala, Alsuhaibani Mohammed, Takanori Maehara, and Ken-ichi
  Kawarabayashi, `Joint word representation learning using a corpus and a
  semantic lexicon', in {\em AAAI'16}, (2016).

\bibitem{bordes2011learning}
Antoine Bordes, Jason Weston, Ronan Collobert, and Yoshua Bengio, `Learning
  structured embeddings of knowledge bases', in {\em AAAI 2011}, (2011).

\bibitem{cer2017semeval}
Daniel Cer, Mona Diab, Eneko Agirre, Inigo Lopez-Gazpio, and Lucia Specia,
  `{S}em{E}val-2017 task 1: Semantic textual similarity multilingual and
  crosslingual focused evaluation', in {\em {S}em{E}val-2017}, (2017).

\bibitem{coates2018frustratingly}
Joshua Coates and Danushka Bollegala, `Frustratingly easy
  meta-embedding--computing meta-embeddings by averaging source word
  embeddings', in {\em NAACL 2018}, (2018).

\bibitem{collobert2008unified}
Ronan Collobert and Jason Weston, `A unified architecture for natural language
  processing: Deep neural networks with multitask learning', in {\em ICML
  2008}, (2008).

\bibitem{conneau2017word}
Alexis Conneau, Guillaume Lample, Marc'Aurelio Ranzato, Ludovic Denoyer, and
  Herv{\'e} J{\'e}gou, `Word translation without parallel data', {\em arXiv
  preprint arXiv:1710.04087}, (2017).

\bibitem{devlin2019bert}
Jacob Devlin, Ming-Wei Chang, Kenton Lee, and Kristina Toutanova, `Bert:
  Pre-training of deep bidirectional transformers for language understanding',
  in {\em ACL 2019}, (2019).

\bibitem{doval:meemiemnlp2018}
Yerai Doval, Jose Camacho-Collados, Luis Espinosa-Anke, and Steven Schockaert,
  `Improving cross-lingual word embeddings by meeting in the middle', in {\em
  EMNLP 2018}, (2018).

\bibitem{faruqui2015retrofitting}
Manaal Faruqui, Jesse Dodge, Sujay~Kumar Jauhar, Chris Dyer, Eduard Hovy, and
  Noah~A Smith, `Retrofitting word vectors to semantic lexicons', in {\em NAACL
  2015}, (2015).

\bibitem{ganitkevitch2013ppdb}
Juri Ganitkevitch, Benjamin Van~Durme, and Chris Callison-Burch, `Ppdb: The
  paraphrase database', in {\em NAACL 2013}, (2013).

\bibitem{ghannay2016word}
Sahar Ghannay, Benoit Favre, Yannick Est{\`e}ve, and Nathalie Camelin, `Word
  embedding evaluation and combination', in {\em LREC 2016}, (2016).

\bibitem{goikoetxea2016single}
Josu Goikoetxea, Eneko Agirre, and Aitor Soroa, `Single or multiple? combining
  word representations independently learned from text and wordnet', in {\em
  AAAI 2016}, (2016).

\bibitem{goikoetxea2015random}
Josu Goikoetxea, Aitor Soroa, and Eneko Agirre, `Random walks and neural
  network language models on knowledge bases', in {\em NAACL 2015}, (2015).

\bibitem{Goikoetxea2018bilingual}
Josu Goikoetxea, Aitor Soroa, and Eneko Agirre, `Bilingual embeddings with
  random walks over multilingual wordnets', {\em Know.-Based Syst.}, (2018).

\bibitem{gouws2015bilbowa}
Stephan Gouws, Yoshua Bengio, and Greg Corrado, `Bilbowa: Fast bilingual
  distributed representations without word alignments', in {\em ICML}, (2015).

\bibitem{halawi2012large}
Guy Halawi, Gideon Dror, Evgeniy Gabrilovich, and Yehuda Koren, `Large-scale
  learning of word relatedness with constraints', in {\em ACM SIGKDD 2012},
  (2012).

\bibitem{heinzerling-strube-2019-sequence}
Benjamin Heinzerling and Michael Strube, `Sequence tagging with contextual and
  non-contextual subword representations: A multilingual evaluation', in {\em
  ACL}, (2019).

\bibitem{hill2014not}
F~Hill, Kyunghyun Cho, Sebastien Jean, C~Devin, and Yoshua Bengio, `Not all
  neural embeddings are born equal', in {\em NIPS 2014 Workshop on Learning
  Semantics}, (2014).

\bibitem{melamud2016context2vec}
Oren Melamud, Jacob Goldberger, and Ido Dagan, `context2vec: Learning generic
  context embedding with bidirectional lstm', in {\em SIGNLL 2016}, (2016).

\bibitem{mikolov2013efficient}
Tomas Mikolov, Kai Chen, Greg Corrado, and Jeffrey Dean, `Efficient estimation
  of word representations in vector space', {\em arXiv preprint
  arXiv:1301.3781}, (2013).

\bibitem{mikolov2018advances}
Tomas Mikolov, Edouard Grave, Piotr Bojanowski, Christian Puhrsch, and Armand
  Joulin, `Advances in pre-training distributed word representations', in {\em
  LREC 2018}, (2018).

\bibitem{mikolov2013exploiting}
Tomas Mikolov, Quoc~V. Le, and Ilya Sutskever, `Exploiting similarities among
  languages for machine translation.', {\em CoRR 2013}, (2013).

\bibitem{mrkvsic2017semantic}
Nikola Mrk{\v{s}}i{\'c}, Ivan Vuli{\'c}, Diarmuid~{\'O} S{\'e}aghdha, Ira
  Leviant, Roi Reichart, Milica Ga{\v{s}}i{\'c}, Anna Korhonen, and Steve
  Young, `Semantic specialization of distributional word vector spaces using
  monolingual and cross-lingual constraints', {\em TACL 2017}, (2017).

\bibitem{scikit-learn}
F.~Pedregosa, G.~Varoquaux, A.~Gramfort, V.~Michel, B.~Thirion, O.~Grisel,
  M.~Blondel, P.~Prettenhofer, R.~Weiss, V.~Dubourg, J.~Vanderplas, A.~Passos,
  D.~Cournapeau, M.~Brucher, M.~Perrot, and E.~Duchesnay, `Scikit-learn:
  Machine learning in {P}ython', {\em Journal of Machine Learning Research},
  {\bf 12},  2825--2830, (2011).

\bibitem{pennington2014glove}
Jeffrey Pennington, Richard Socher, and Christopher Manning, `{G}love: Global
  vectors for word representation', in {\em EMNLP 2014}, (2014).

\bibitem{Peters:2018}
Matthew~E. Peters, Mark Neumann, Mohit Iyyer, Matt Gardner, Christopher Clark,
  Kenton Lee, and Luke Zettlemoyer, `Deep contextualized word representations',
  in {\em NAACL 2018}, (2018).

\bibitem{raunak2017simple}
Vikas Raunak, `Simple and effective dimensionality reduction for word
  embeddings', in {\em "Learning with Limited Labeled Data: Weak Supervision
  and Beyond" workshop at NIPS}, (2017).

\bibitem{salle2016matrix}
Alexandre Salle, Aline Villavicencio, and Marco Idiart, `Matrix factorization
  using window sampling and negative sampling for improved word
  representations', in {\em ACL 2016}, (2016).

\bibitem{shao2017hcti}
Yang Shao, `Hcti at semeval-2017 task 1: Use convolutional neural network to
  evaluate semantic textual similarity', in {\em SemEval-2017}, (2017).

\bibitem{socher2011semi}
Richard Socher, Jeffrey Pennington, Eric~H Huang, Andrew~Y Ng, and
  Christopher~D Manning, `Semi-supervised recursive autoencoders for predicting
  sentiment distributions', in {\em EMNLP}, (2011).

\bibitem{speer2017conceptnet}
Robyn Speer, Joshua Chin, and Catherine Havasi, `Conceptnet 5.5: An open
  multilingual graph of general knowledge', in {\em AAAI 2017}, (2017).

\bibitem{speer2017semeval}
Robyn Speer and Joanna Lowry-Duda, `Conceptnet at semeval-2017 task 2:
  Extending word embeddings with multilingual relational knowledge', in {\em
  SemEval-2017}, (2017).

\bibitem{Fei:Learning}
Fei Sun, Jiafeng Guo, Yanyan Lan, Jun Xu, and Xueqi Cheng, `Learning word
  representations by jointly modeling syntagmatic and paradigmatic relations',
  in {\em ACL 2015}, (2015).

\bibitem{turian2010word}
Joseph Turian, Lev Ratinov, and Yoshua Bengio, `Word representations: a simple
  and general method for semi-supervised learning', in {\em ACL 2010}, (2010).

\bibitem{vulic2016role}
Ivan Vuli{\'c} and Anna Korhonen, `On the role of seed lexicons in learning
  bilingual word embeddings', in {\em ACL}, (2016).

\bibitem{wieting2015paraphrase}
John Wieting, Mohit Bansal, Kevin Gimpel, and Karen Livescu, `From paraphrase
  database to compositional paraphrase model and back', {\em TACL 2015},
  (2015).

\bibitem{yin-schutze:2016:P16-1}
Wenpeng Yin and Hinrich Sch\"{u}tze, `Learning word meta-embeddings', in {\em
  ACL 2016}, (2016).

\end{thebibliography}
\end{document}